\documentclass[a4paper]{article}

\usepackage{INTERSPEECH2015}

\usepackage{graphicx}
\usepackage{amssymb,amsmath,bm}
\usepackage{textcomp}
\usepackage{multirow}
\usepackage{arydshln}

\newcommand{\phat}[1]{\hat{#1}}

\ninept

\title{Finding phonemes: improving machine lip-reading}

\makeatletter
\def\name#1{\gdef\@name{#1\\}}
\makeatother \name{{\em Helen L. Bear$^1$, Richard W. Harvey$^1$, Yuxuan Lan$^1$}}

\address{$^1$University of East Anglia, UK\\
  {\small \tt \{helen.bear,r.w.harvey,y.lan\}@uea.ac.uk}
}

\begin{document}

  \maketitle
  \begin{abstract}
In machine lip-reading there is continued debate and research around the correct classes to be used for recognition. 

In this paper we use a structured approach for devising speaker-dependent viseme classes, which enables the creation of a set of phoneme-to-viseme maps where each has a different quantity of visemes ranging from two to 45. Viseme classes are based upon the mapping of articulated phonemes, which have been confused during phoneme recognition, into viseme groups.

Using these maps, with the LiLIR dataset, we show the effect of changing the viseme map size in speaker-dependent machine lip-reading, measured by word recognition correctness and so demonstrate that word recognition with phoneme classifiers is not just possible, but often better than word recognition with viseme classifiers. Furthermore, there are intermediate units between visemes and phonemes which are better still. 

\end{abstract}

  \section{Introduction}

Although visemes are yet to be formally defined, many possibilities can be found across literature \cite{bear2014phoneme, chen1998audio, fisher1968confusions, Hazen1027972}. Here we use the definition ``a viseme is a visual cue representative of a subset of phonemes on the lips". Therefore, a set of viseme classifiers is inherently smaller than a set of phoneme classifiers. Whilst this means that there are more training samples per class (addressing the limitation of currently available dataset sizes), this also introduces generalisation between articulated sounds. So, to find optimal viseme classes, we need to minimise this generalisation in order to maximise recognition of correct utterances, but also maximise the use of the data available. 

The relationship between phonemes (the units of acoustic speech) and visemes (the units of visual speech) can be described with Phoneme-to-Viseme (P2V) maps. In \cite{bear2014phoneme} it is shown how these maps can be derived automatically from phoneme confusions.  A by-product of the method is that we can control how many visemes we need. This allows considerable precision when answering questions about the optimal number and nature of visemes. 
 \section{Data}
 
Our selected dataset is LiLIR \cite{lan2010improving}. This data consists of 12 British speakers (seven male and five female), 200 utterances per speaker of resource management context independent sentences from \cite{fisher1986darpa} which totals around 1000 words. The original videos were recorded in high definition and in a full-frontal position. Individual speakers are tracked using Active Appearance Models \cite{Matthews_Baker_2004} and we extract features of concatenated shape and appearance information. 

The pronunciation dictionary used throughout these experiments is British English \cite{beep} which we take to be represented by 46 phonemes.
  
  \section{Method}

\begin{figure*}[t]
        \centering
        \includegraphics[width=\linewidth]{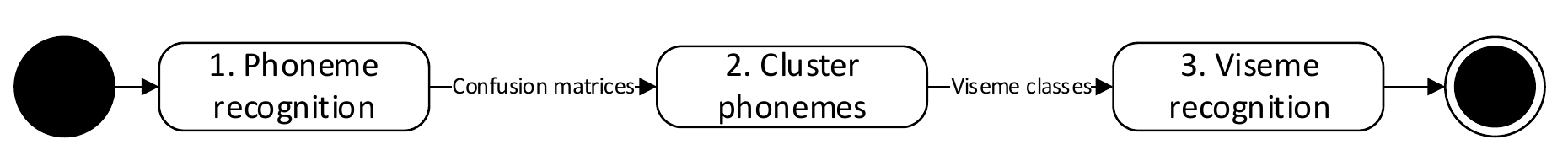}
        \caption{{\it Three step process for word recognition from visemes.}}
        \label{fig:process}
      \end{figure*}

A high level overview of our method is shown in Figure~\ref{fig:process} and is described in \cite{bear2014phoneme}. We begin by performing word recognition using classifiers based upon phoneme labels. This provides us with both a baseline to benchmark against and, crucially, a set of confusion matrices for each speaker which are used to cluster together potential monophones. 

However, we undertake a different clustering process (section~\ref{sec:two}) during which we make a new P2V mapping each time a phoneme is re-classified to a new viseme grouping, thereby deriving up to 45 (subject to the number of phonemes recognised during the phoneme recognition stage) P2V maps per speaker. These new classifiers (visemes) are then used to repeat our word recognition task. 

We use the word recognition as our performance measure as this normalises for variance in training samples for each set of classifiers. We note that it is not the performance itself which is relevant here, rather it is any improvement a variance in classes can provide. The reader should also note that we are not suggesting our clustering process will deliver the optimum visemes but rather address our need in this case for a method to enable a controlled comparison of the visemes.

 \subsection{Step one: phoneme recognition}
 \label{sec:one}
 
We implement 10-fold cross-validation with replacement \cite{efron1983leisurely}, of 200 sentences per speaker, 20 are randomly selected as test samples and these are not included in the training folds. Using the HTK toolkit \cite{htk34} to use Hidden Markov Model (HMM) classes, we flat-start the HMMs, re-estimate them 11 times with forced alignment between seventh and eighth estimates.  Our prototype is based upon a Gaussian mixture of five components and three state HMMs. We use a single-state tied short-pause, or `sp' HMM for short silences between words in the sentence utterances. We also use a bigram word network to support recognition. There are a maximum of 46 phonemes within our phoneme recognition results, but not all speakers used all phonemes within their speech utterances. 
 
 \subsection{Step two: speaker-dependent phoneme clustering}
 \label{sec:two}

We cluster the phonemes into new visemes classes as follows; we have 10 confusion matrices for each speaker (one from each fold), these are summed together to form one confusion matrix representing all confusions for that speaker. We start with this phoneme confusion matrix: 

\begin{equation}
	[K_{m}]_{ij} = N (\phat{p}_j | p_i)\quad
        	\label{eq3}
\end{equation}

where the $ij^{th}$ element is the count of the number of times phoneme $i$ is classified as phoneme $j$. Our algorithm works with the column normalised version, 

\begin{equation}
	[P_m]_{ij} = Pr\{p_i | \phat{p}_j \} \quad
        	\label{eq4}
\end{equation}

the probability that, given a classification of $p_j$ that the phoneme really was $p_i$. The subscript $m$ in $K_m$ and $P_m$ indicates that  $K_m$ and $P_m$ have $m^{2}$ elements ($m$ phonemes). We merge phonemes by looking for the two most confused phonemes and hence create a new class with confusions $K_{m-1}, P_{m-1}$.

Specifically for each possible merged pair, $Pr,Ps$, we calculate a score:
\begin{equation}
	q = [P_{m}]_{rs} + [P_{m}]_{sr} \quad \\
	= Pr\{\phat{P}r | Ps \} + Pr\{Pr | \phat{P}s\}
        	\label{eq5}
\end{equation}

Phonemes are assigned to one of two classes, $V \& C$, vowels and consonants. Vowels and consonants can not be mixed. The pair with the highest $q$ is merged. Equal scores are broken randomly. This process is repeated until  $M = 2$. Each intermittent step, $M = 45,44,43 ...  2$ forms a possible set of visual units.  

This is a more formal approach than used in \cite{bear2014phoneme} and incorporates their conclusions that vowel and consonant phonemes should not be clustered together when devising phoneme-to-viseme mappings. An example P2V mapping is shown in Table~\ref{tab:example}. 

\begin{table}
\centering
\begin{tabular}{|l|l|}
\hline
Viseme & Phonemes \\
\hline \hline
V01 & /ax/ \\
V02 & /v/ \\
V03 & /oy/ \\
V04 & /f/ /zh/ /w/ \\
V05 & /k/ /b/ /d/ /th/ /p/ \\
V06 & /l/ /jh/ \\
V07 & /g/ /m/ /z/ /y/ /ch/ /dh/ /s/ /r/ /t/ /sh/ \\
V08 & /n/ /hh/ /ng/ \\
V09 & /ea/ /ae/ /ao/ /uw/ /oh/ /ia/ /ey/ /ua/ /er/ \\
V10 & /ay/ /aa/ /ah/ /aw/ /uh/ /ow/ /ih/ /iy/ /az/ /eh/ \\
\hline
\end{tabular}
\caption{An example P2V map, this is the P2V for Speaker 01 with ten visemes}
\label{tab:example}
\end{table}

\subsection{Step three: viseme recognition}
 \label{sec:visRecog}

Similar to Step one, we implement 10-fold cross-validation with replacement \cite{efron1983leisurely}, of 200 sentences per speaker, 20 are randomly selected as test samples and these are not included in the training folds. Using the HTK toolkit \cite{htk34} to use Hidden Markov Model (HMM) classes, we flat-start the HMMs, re-estimate them 16 times over with forced alignment between seventh and eighth estimates.

Our prototype is based upon a gaussian mixture of five components and three state HMMs. We use a single-state tied short-pause, or `sp' HMM for short silences between words in the sentence utterances. We also use a bigram word network to support recognition, apply a grammar scale factor of $1.0$ (shown to be optimum in Howell's thesis~\cite{howellPhD}) and apply a transition penalty of $0.5$.

This time around we have viseme classes to use as recognizers. By using these sets of classes which have shown in step one are confusing on the lips, we perform recognition for each class set. In total this is 45, where the smallest set is of two classes (one with all the vowel phonemes and the other all the consonant phonemes), and the largest set is of 45 classes with one phoneme in each - a repeat of the phoneme recognition task but using only phonemes which we know to have been identifiable. 

\section{Discussion}
We note that word recognition performance of the HMMs can be measured by both correctness, $C$, and accuracy, $A$, of the recognition classes,
\begin{equation}
	C = \displaystyle \frac{N-D-S}{N},\quad
        	\label{eq1}
\end{equation}
\begin{equation}	
	A = \displaystyle \frac{C-I}{N},
	\label{eq2}
\end{equation}
where $S$ is the number of substitution errors, $D$ is the number of deletion errors, $I$ is the number of insertion errors and $N$ the total number of labels in the reference transcriptions~\cite{htk34}.

\begin{figure*} [!pht]
\centering
\setlength{\tabcolsep}{1pt}
\begin{tabular} {c c}
	\includegraphics[width=0.5\textwidth]{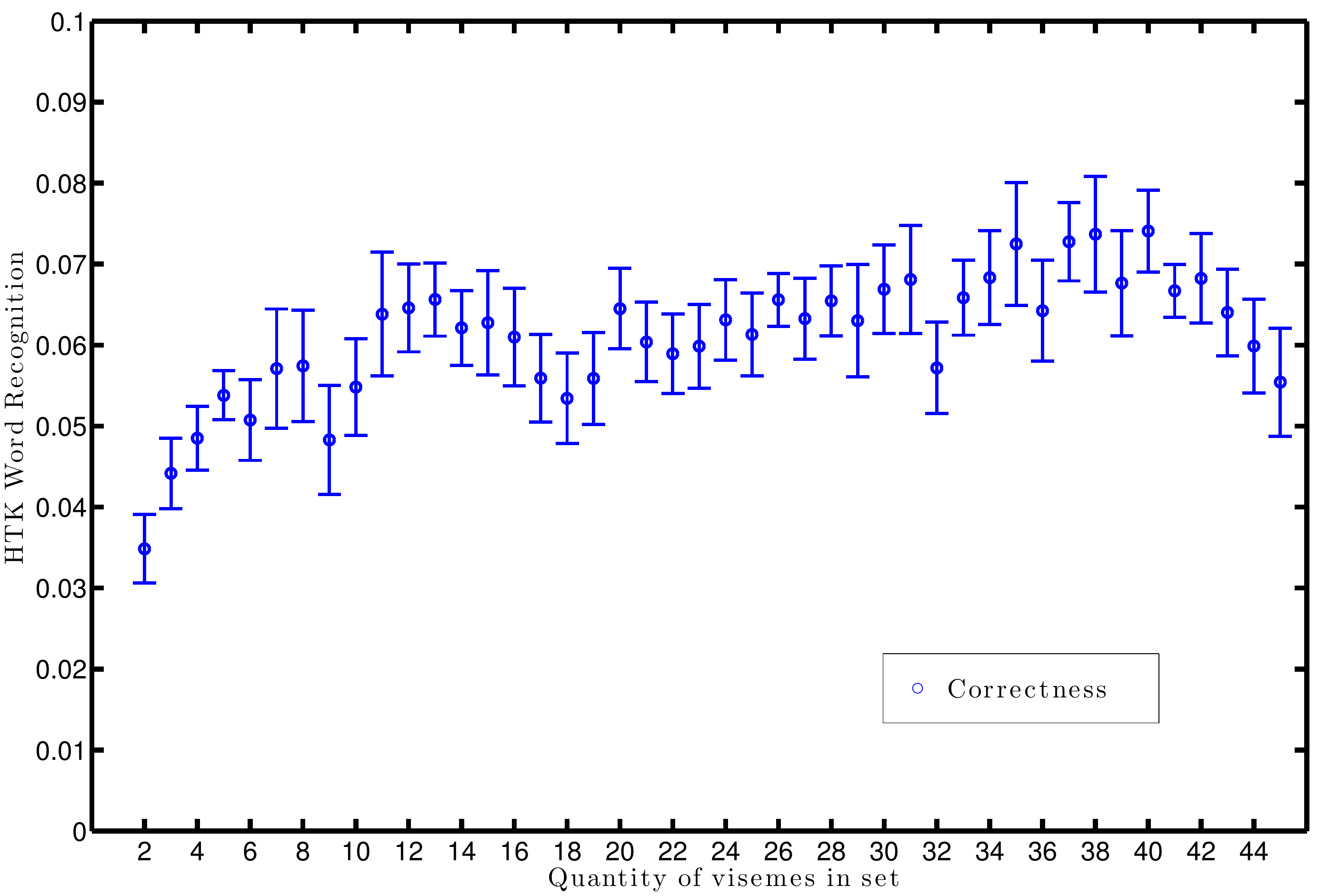} & \includegraphics[width=0.5\textwidth]{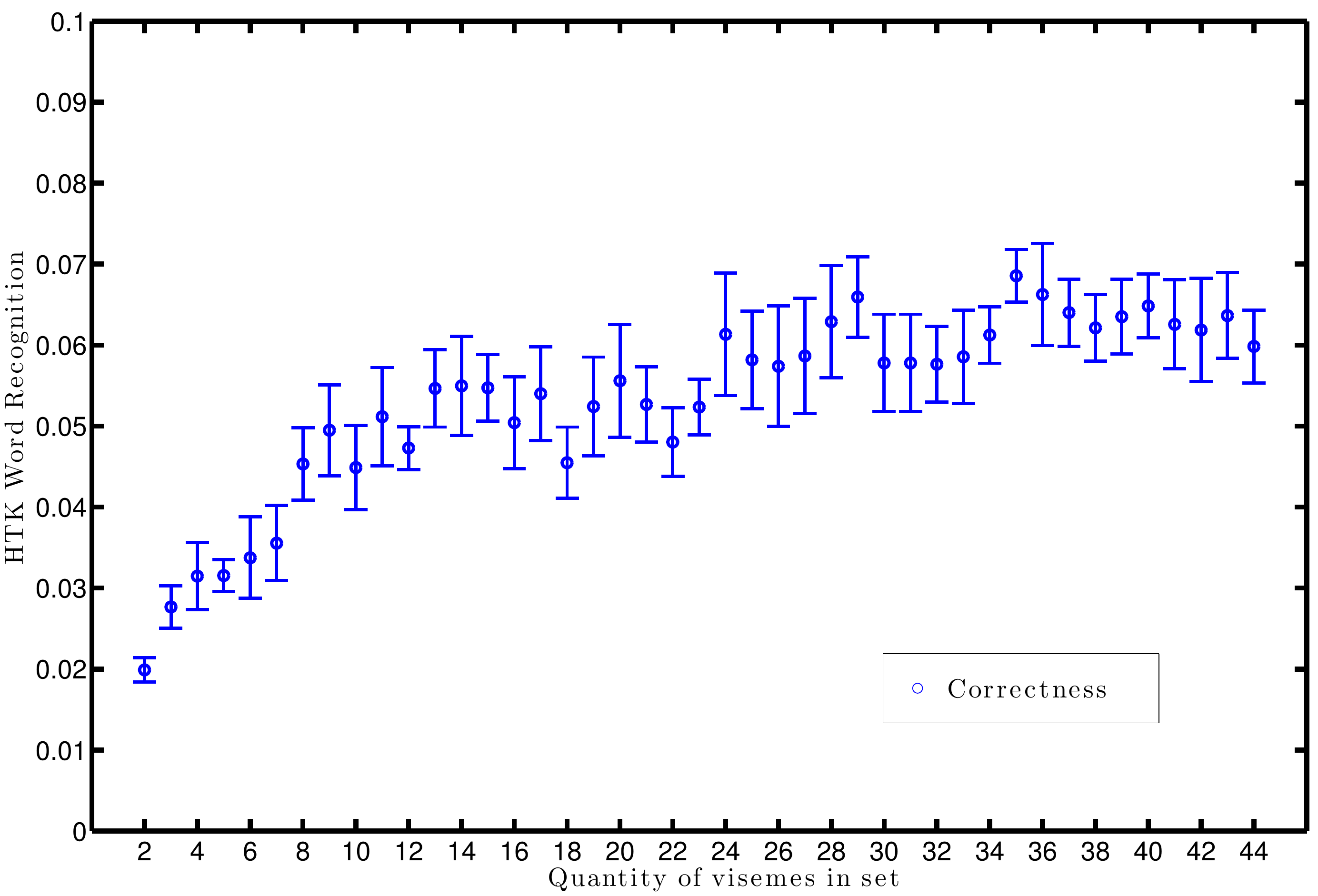} \\
	(g) Speaker 7 & (h) Speaker  8 \\
	\includegraphics[width=0.5\textwidth]{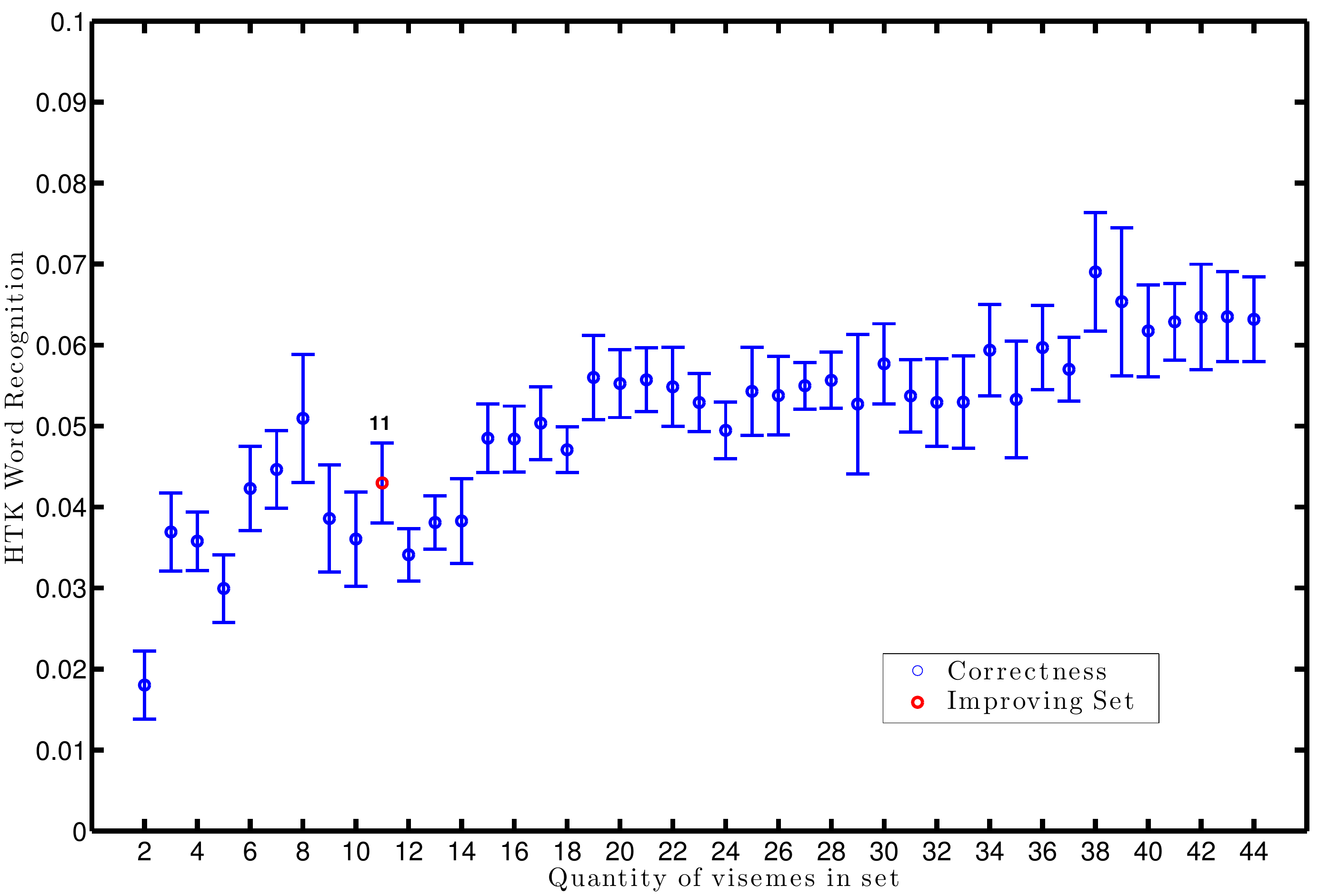} & \includegraphics[width=0.5\textwidth]{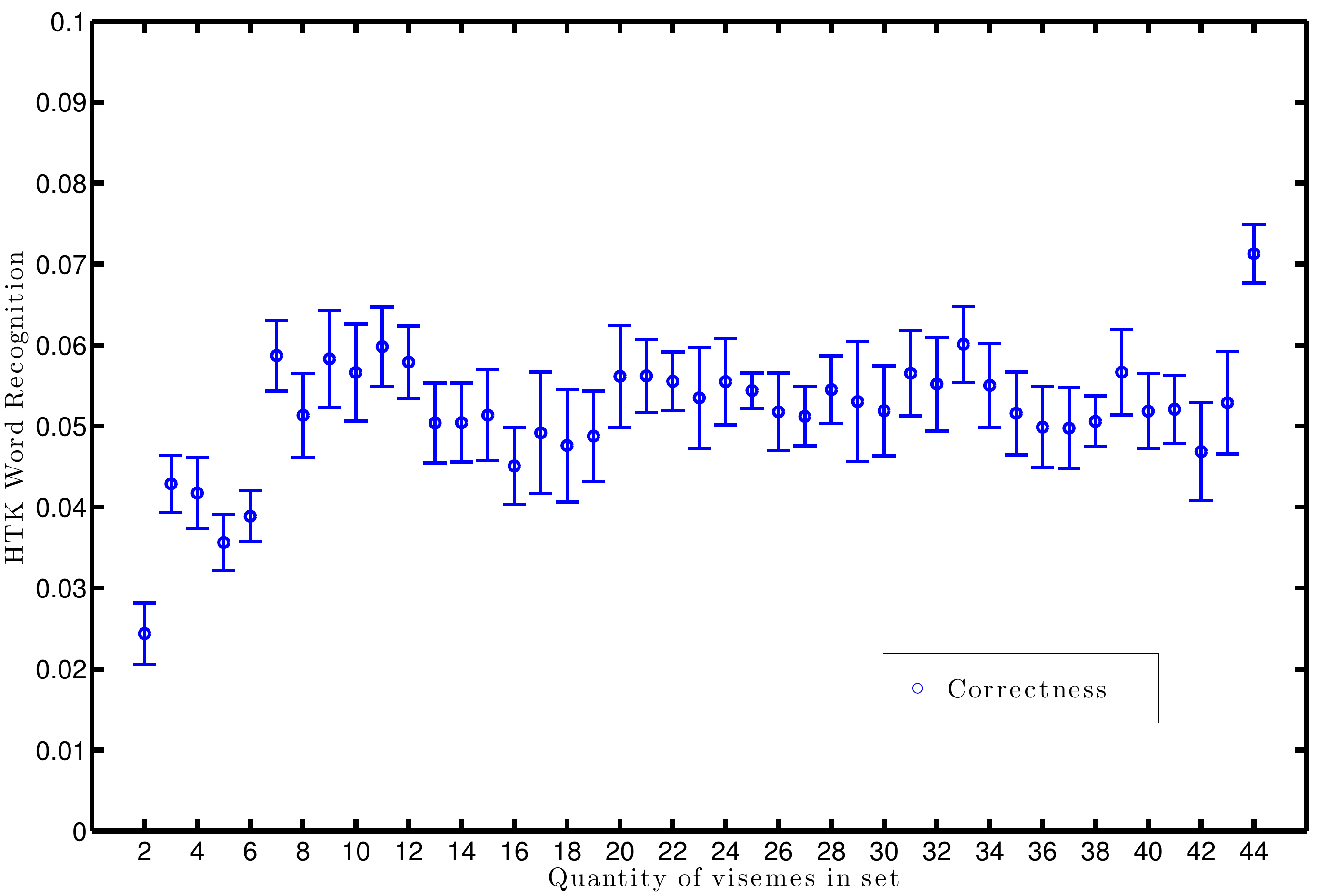} \\
	(i) Speaker 9 & (j) Speaker 10 \\
	\includegraphics[width=0.5\textwidth]{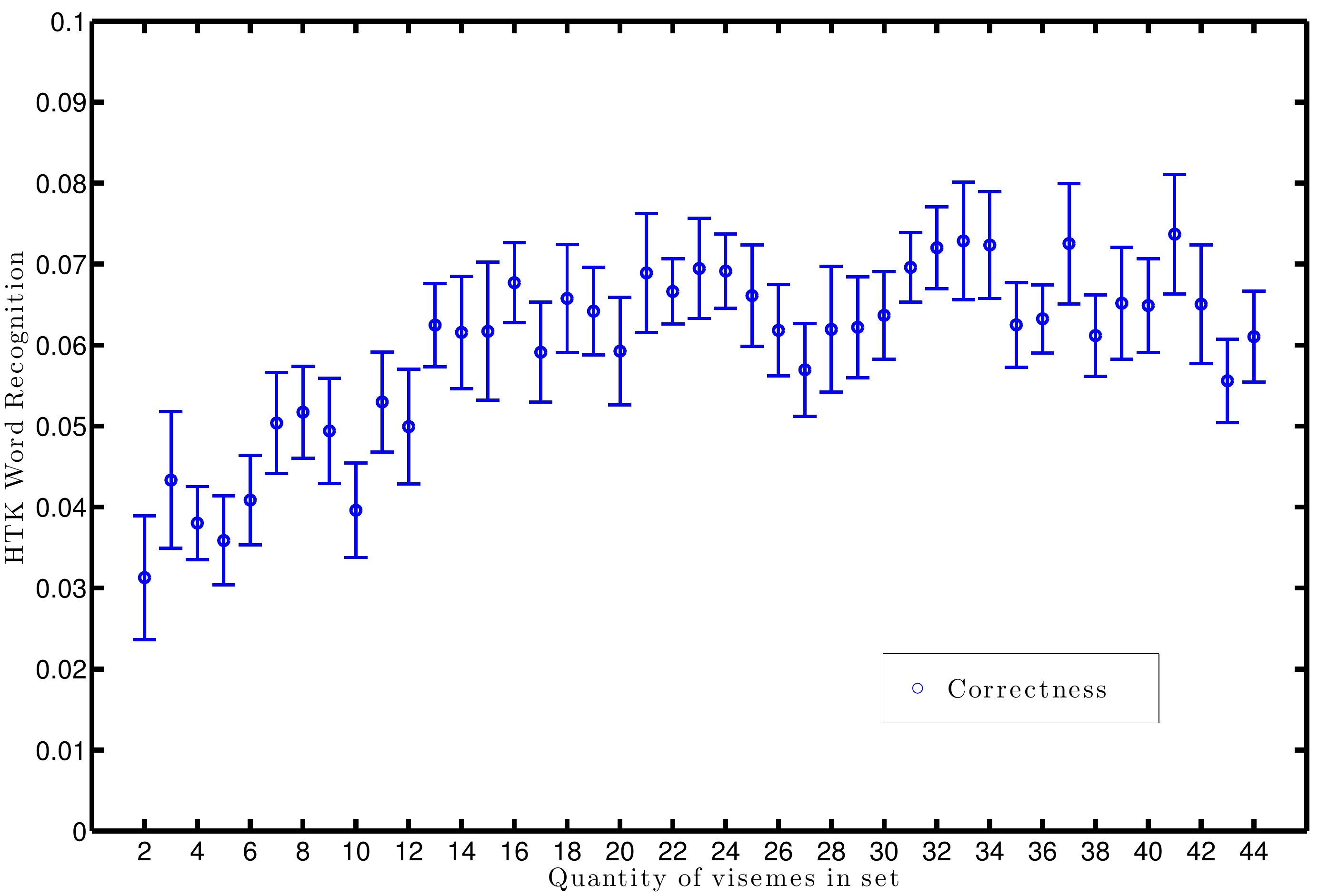} & \includegraphics[width=0.5\textwidth]{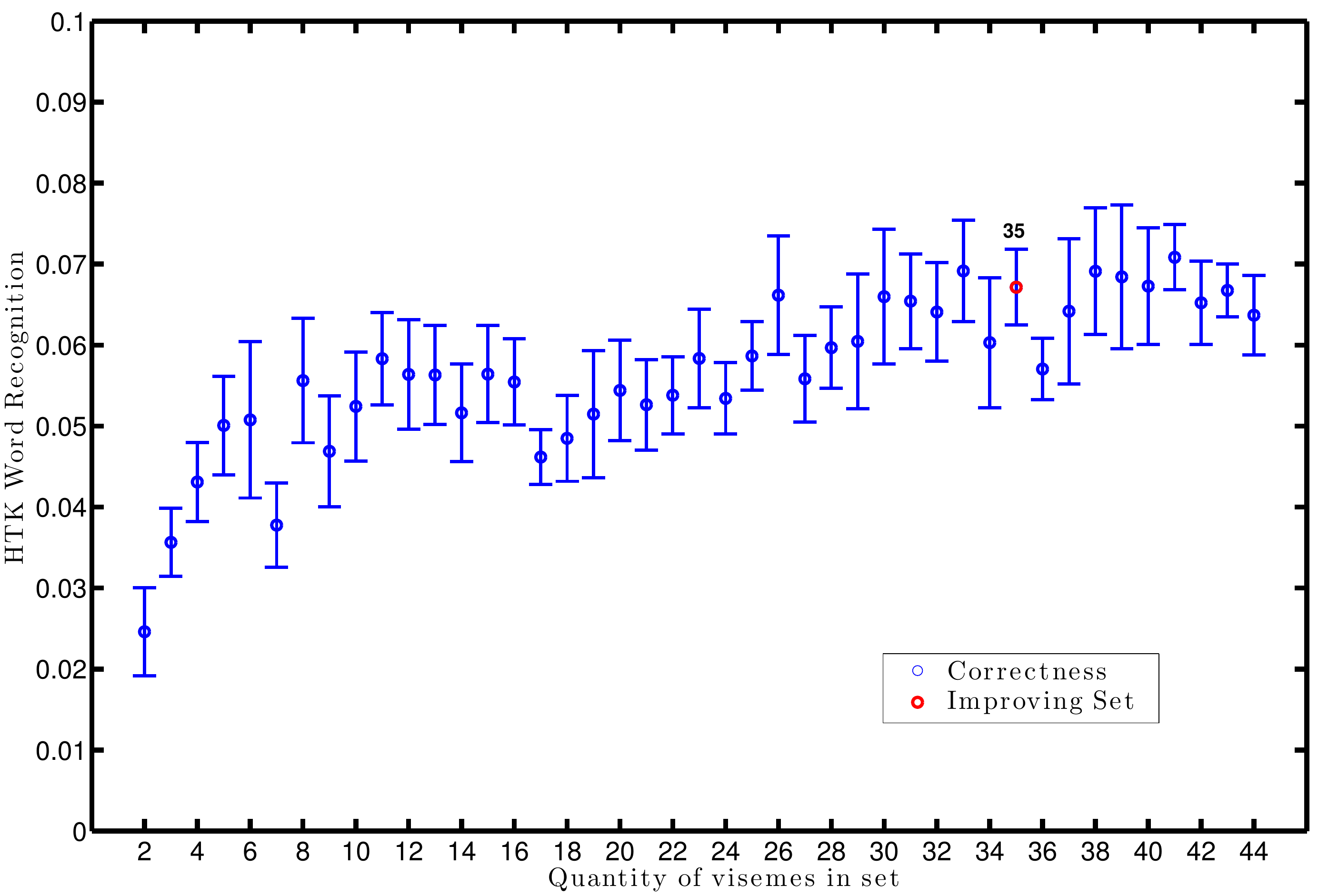} \\
	(k) Speaker 11 & (l) Speaker 12 \\
\end{tabular}
\caption{\label{tab:indSpeakerFigs} {\it Individual speaker word recognition in correctness $C$ for all viseme map sizes}}
\end{figure*}

Figure~\ref{tab:indSpeakerFigs} (subfigures a-l), show the correctness for all 12 speakers. Viseme sets containing fewer visemes produce more viseme strings that represent more than one word: homophones. An example of a homophone in these data are the words `port' and `bass'. Using Speaker 1's 10-viseme P2V map these both become `v5 v9 v7'  i.e. a single identifier for identifying two words. Thus distinguishing between `port' and `bass' becomes impossible. The effect of these can be seen on the left side of the graphs in Figure~\ref{tab:indSpeakerFigs}.

Although the correctness scores are low they are all significantly above chance. The results for each speaker vary but the overall trend is very clear. Superior performances are to be found with larger numbers of visemes. Note that, had we reported viseme error (as is commonplace) then this effect is not visible and the imperative for large numbers of visemes would be missed. 

Also in Figure~\ref{tab:indSpeakerFigs} (subfigures a-l), class sets are highlighted in red and labelled which show where a particular combination of two previous viseme classes delivers a significant improvement in recognition. These combinations are listed in Table~\ref{tab:merges}. Whilst there is no apparent pattern through these pairings, this does further reinforce our knowledge that all speakers are visually unique and how difficult finding a set of cross-talker viseme sets will be when different phonemes require alternative grouping arrangements for each individual.

\begin{figure*} [!pht]
\centering
\setlength{\tabcolsep}{1pt}
\begin{tabular} {c c}
	\includegraphics[width=0.5\textwidth]{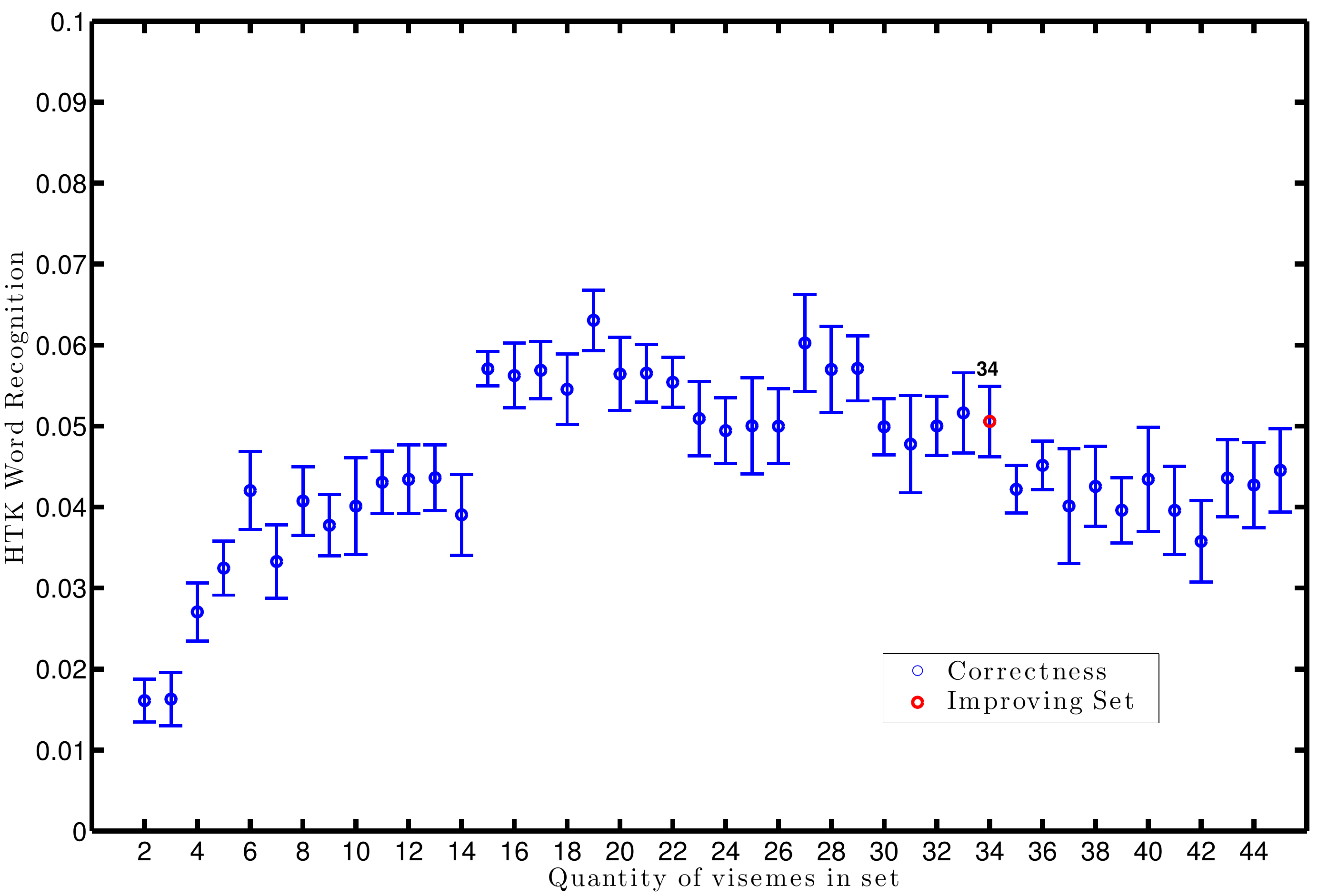} & \includegraphics[width=0.5\textwidth]{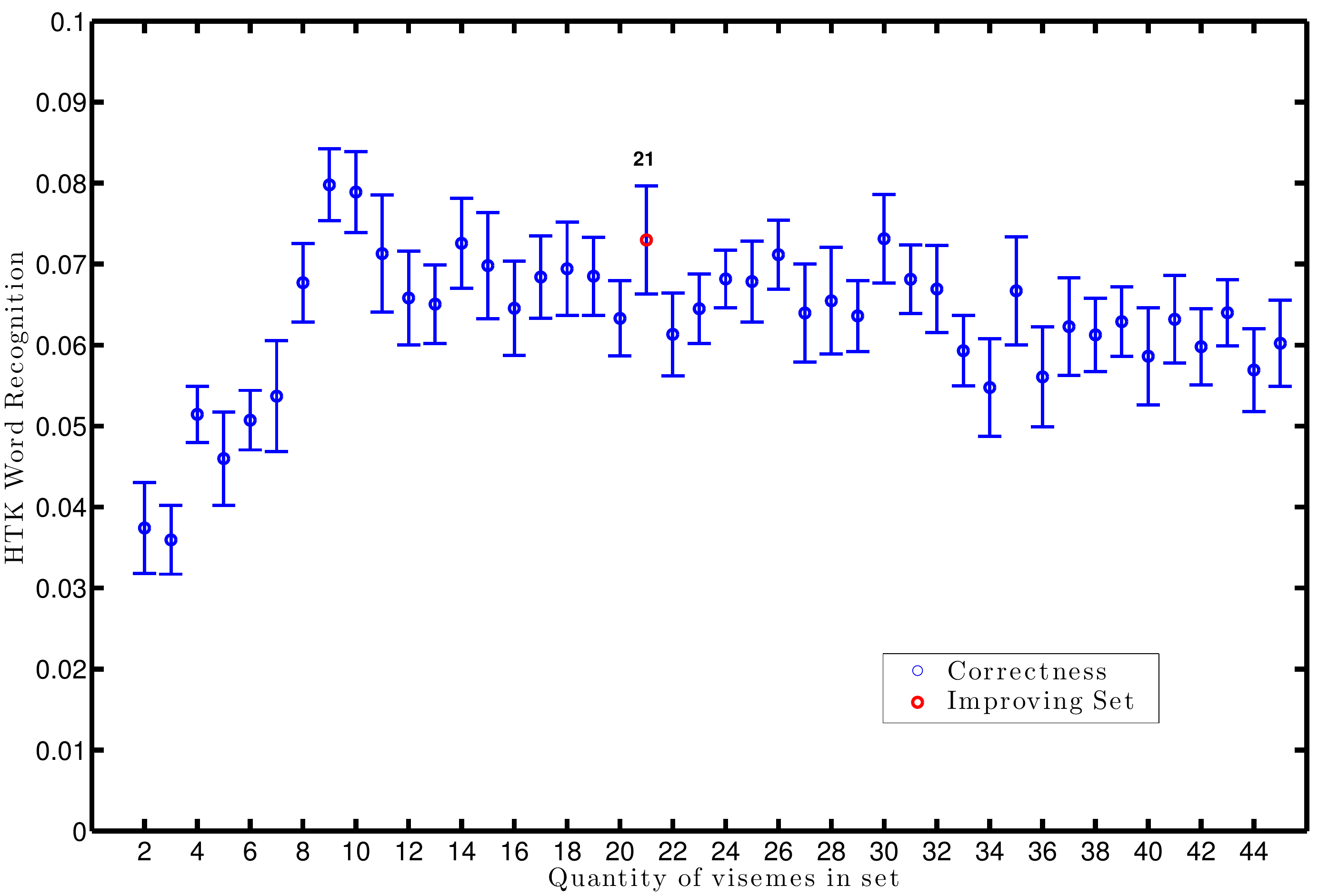} \\
	(a) Speaker 1 & (b) Speaker 2 \\
	\includegraphics[width=0.5\textwidth]{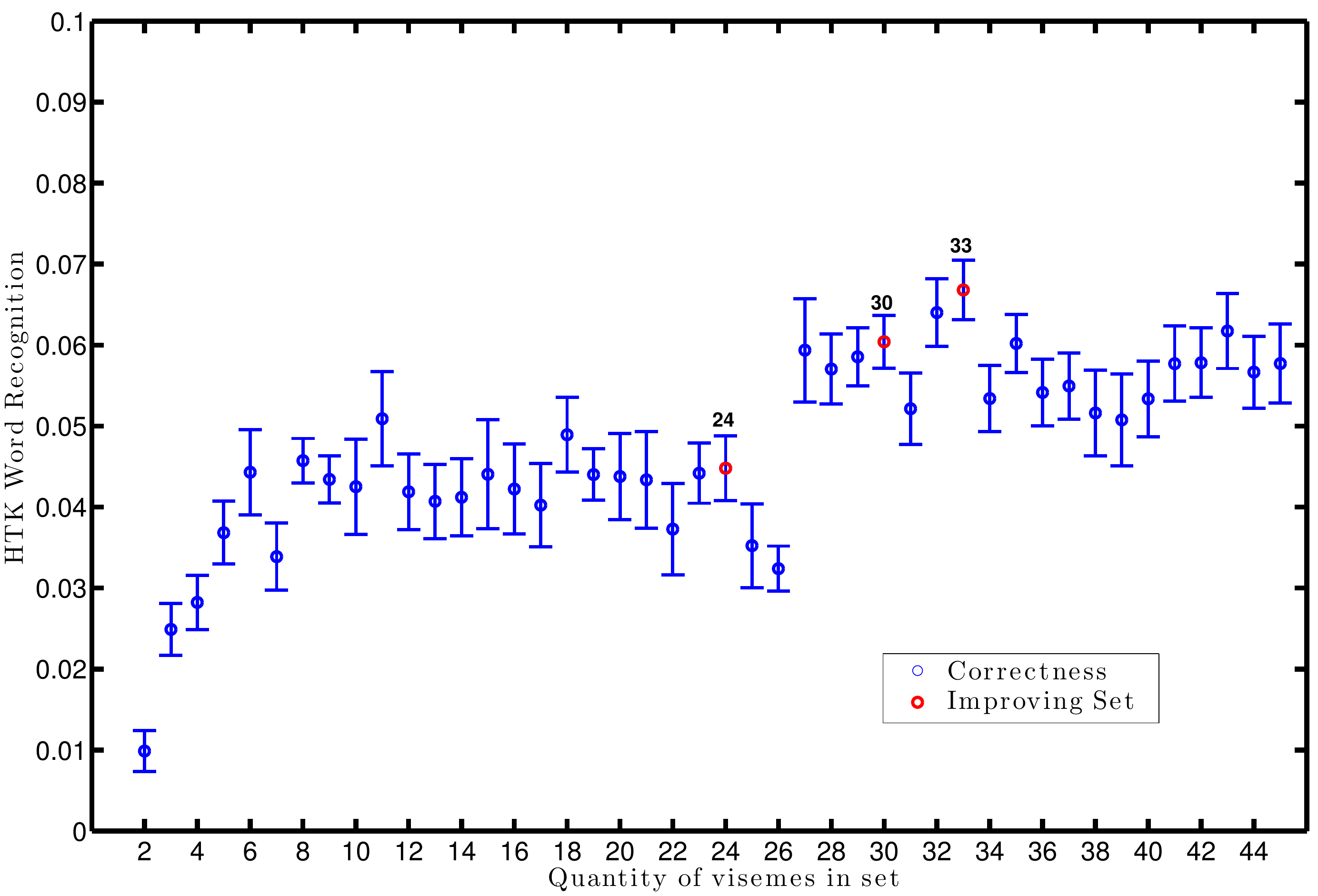} & \includegraphics[width=0.5\textwidth]{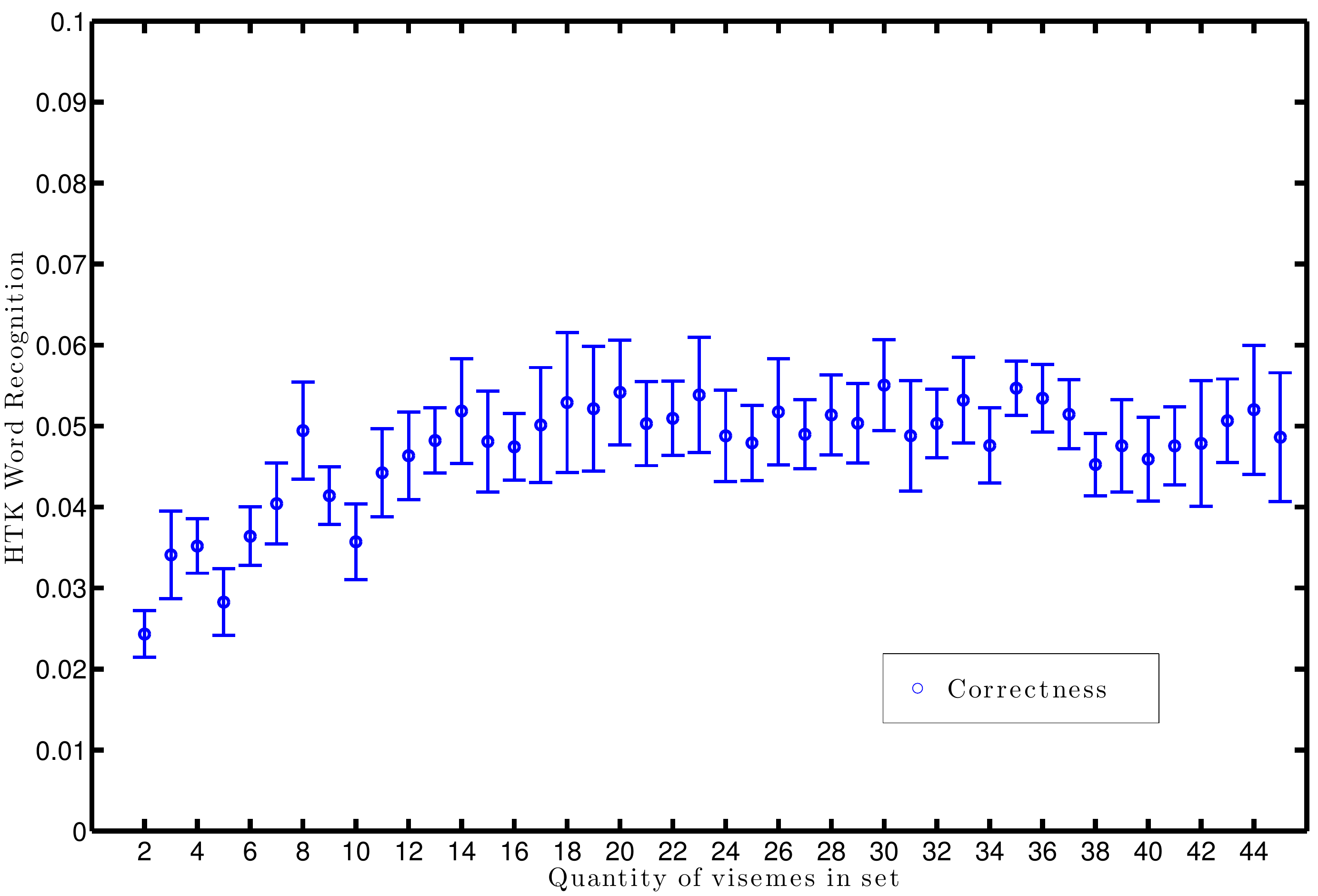} \\
	(c) Speaker 3 & (d) Speaker 4 \\
	\includegraphics[width=0.5\textwidth]{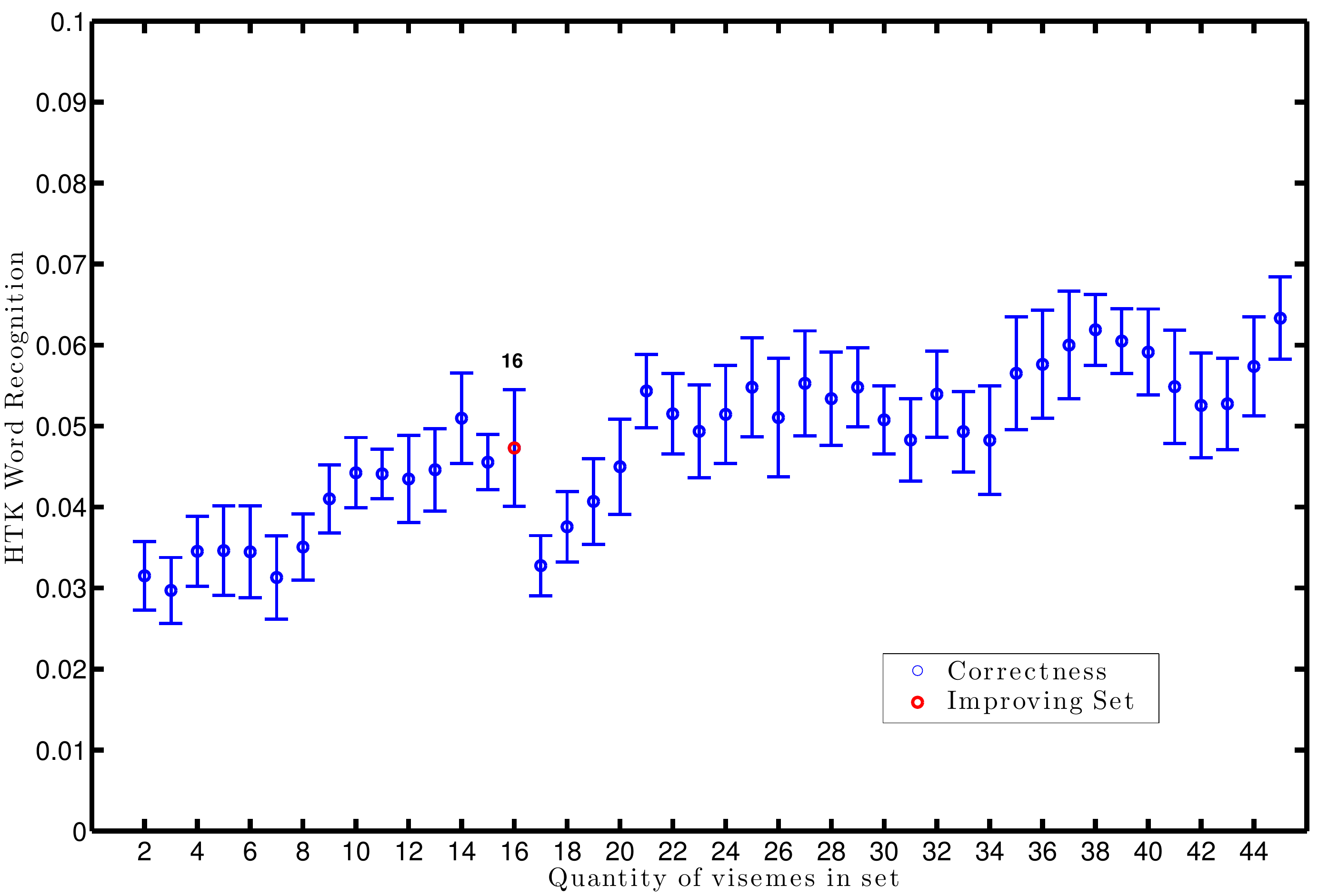} & \includegraphics[width=0.5\textwidth]{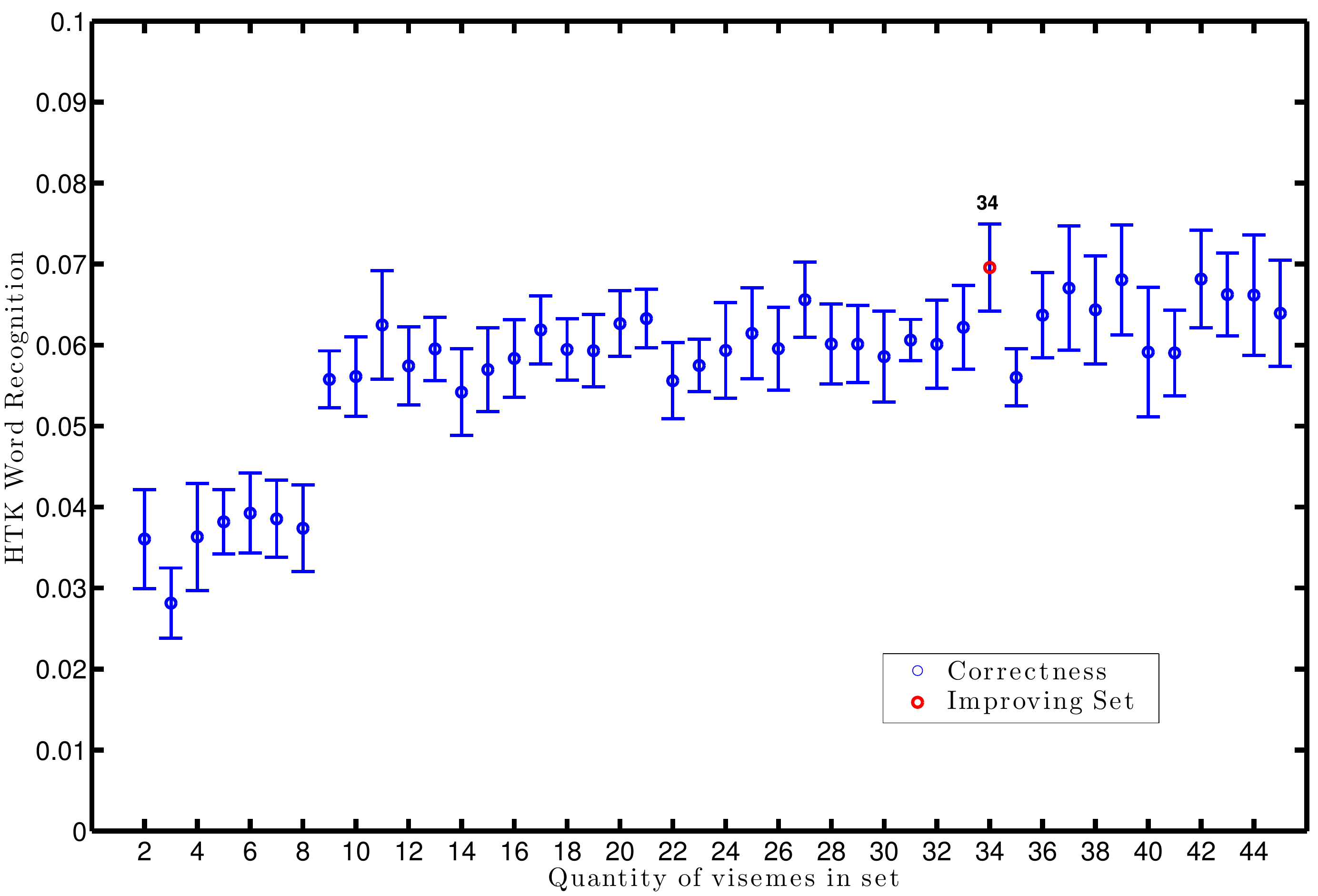} \\
	(e) Speaker 5 & (f) Speaker 6\\
\end{tabular}
\end{figure*}

\begin{table*}[!ht]
\centering
\begin{tabular}{| l || l | l | l || l | l |}
\hline
Speaker & Set No & $V_i$ & $V_j$ & Set No & $V_n$ \\
\hline \hline
SP01 & 35 & /s/ /r/ & /dh/ & 34 & /s/ /r/ /dh/  \\
SP02 & 22 & /d/ & /z/ /y/ & 21 & /d/ /z/ /y/ \\
SP03 & 34 & /b/ /ch/ & /zh/ & 33 & /b/ /ch/ /zh/ \\
SP03 & 31 & /zh/ /b/ /ch/ & /z/ & 30 & /zh/ /b/ /ch/ /z/ \\
SP03 & 25 & /p/ /r/ & /ng/ & 24 & /p/ /r/ /ng/ \\
SP05 & 17 & /ae/ & /eh/ & 16 & /ae/ /eh/ \\ 
SP06 & 35 & /ae/ /ah/ & /iy/ & 34 & /ae/ /ah/ /iy/ \\
SP09 & 12 & /b/ /w/ /v/ & /jh/ /hh/ & 11 & /b/ /w/ /v/ /jh/ /hh/ \\
SP12 & 36 & /ah/ & /ao/ & 34 & /ah/ /ao/ \\ 
\hline 
\end{tabular}
\caption{Viseme class merges which improve word recognition}
\label{tab:merges}
\end{table*}

\begin{table*}[!ht]
\centering
\begin{tabular}{|l|r|r|r|r|r|r|r|r|r|r|r|r|}
	\hline
	Speaker & 1 & 2 & 3 & 4 & 5 & 6 & 7 & 8 & 9 & 10 & 11 & 12 \\
	Phoneme $C$ & 0.045 & 0.060 & 0.058 & 0.049 & 0.063 & 0.063 & 0.055 & 0.090 & 0.063 & 0.071 & 0.061 & 0.064 \\ 
	\hline
\end{tabular}
\caption{Phoneme correctness values for each speaker, these are on the right hand side of each respective subfigure in Figure~\ref{tab:indSpeakerFigs}}
\label{tab:pr_vals}
\end{table*}

 \begin{figure*}[!ht]
        \centering
        \includegraphics[width=0.8\linewidth]{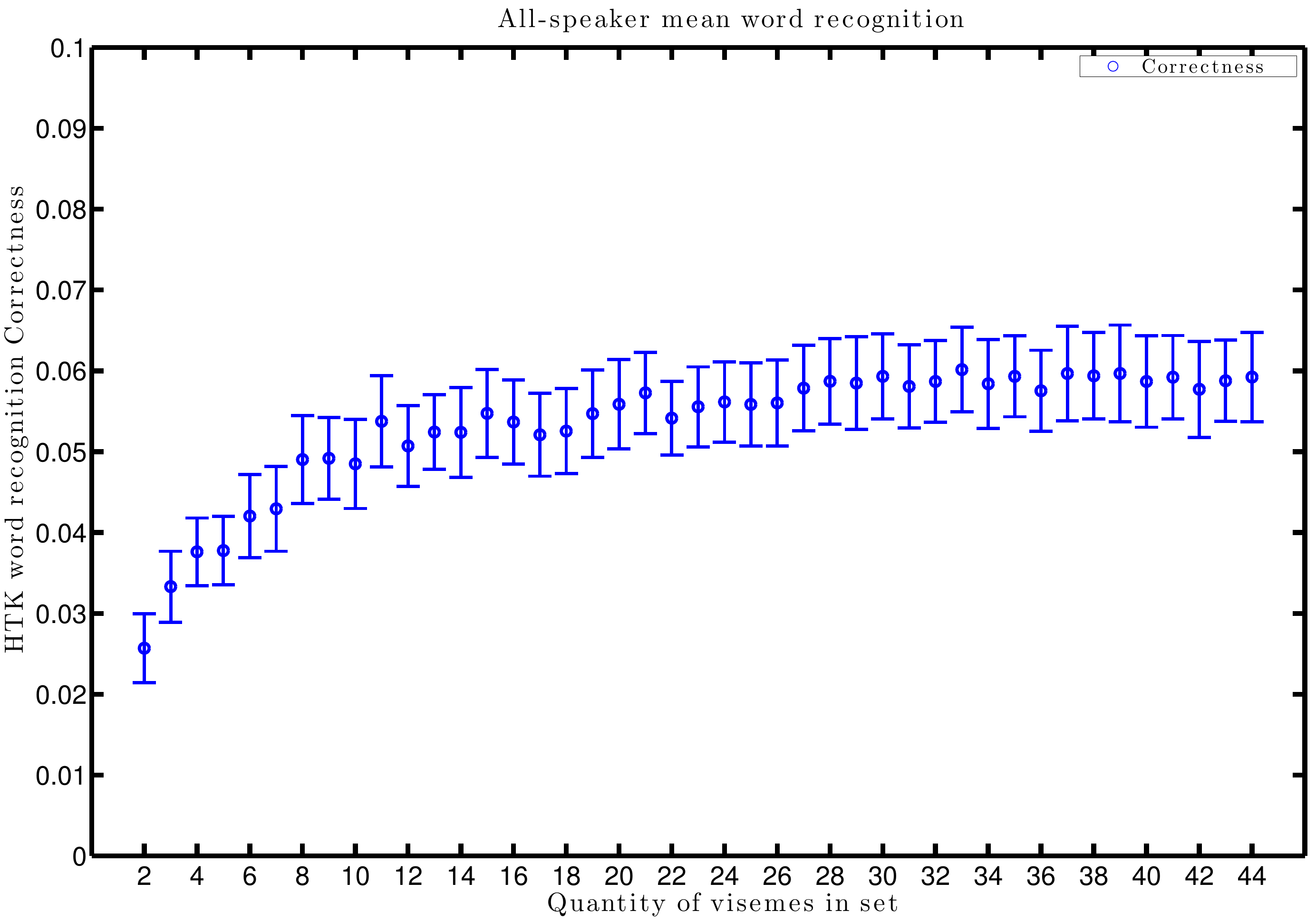}
        \caption{{\it Word recognition measured by correctness of the classifiers.  Error bars show $\pm$ one standard error.}}
        \label{fig:correctness}
      \end{figure*}
     
     As has been noted before \cite{hazen2006visual} the conventional wisdom which is that visemes are needed for lip-reading is not bourne out by these experiments. However it is an over simplification to assert that better lip-reading can be achieved with phonemes than visemes. It is true that, generally speaking, larger numbers of visemes out-perform smaller numbers, but the curves in Figure~\ref{tab:indSpeakerFigs} are far from monotonic. Even Figure~\ref{fig:correctness}, which is the mean performance over all speakers, is not monotonic. 

There are a number of proposed phoneme-to-viseme maps in the literature, typically they generate between 10 and 20 visemes (see \cite{bear2014phoneme} for a summary) - the well known Lee set has six consonant visemes and five vowels \cite{lee2002audio}; Jeffers eight \& three  \cite{jeffers1971speechreading} and so on. Looking at Figures~\ref{tab:indSpeakerFigs} \&~\ref{fig:correctness} there is certainly a rapid drop-off in performance for fewer than ten visemes but the region between ten and 20 contains the optimum viseme set for three out of the 12 speakers which is no more than chance. In other words, for each speaker there is an optimal number of visual units (shown by the best performing result in Figure~\ref{tab:indSpeakerFigs}) but that optimal number is not related to any of the conventional viseme definitions, nor is the number of phonemes. The correctness of the phoneme recognition for each speaker is shown in Table~\ref{tab:pr_vals}.

The two factors at play in these graphs are, the underlying accuracy with which the visual units represent the mouth shape and appearances versus the introduction of homophones. For large numbers of visemes we are close to phonetic recognition, (with fewer homophones) but we run the risk of visual units which are not visually very distinctive - several of the HMM models will ``match" on a particular sub-sequence. This latter problem creates a decoding lattice in which there are several near equal probability paths which, in turn, implies that state-of-the-art language models would improve results still further. 
 
 \section{Conclusions}
 We have described a method that allows us to construct any number of visual units. We remind the reader that we are not proposing that our visemes are the best, our priority in this case is a method for enabling comparison of viseme sets in a controlled manner. 
 
 The presence of an optimum is a result of two competing effects. In the first, as the number of visemes shrinks the number of homophones rises and it becomes more difficult to recognise words (correctness drops). In the second, as the number of visemes rises we run out of training data to learn the subtle differences in lip-shapes (if they exist), so again, correctness drops.
 
 Thus, the optimum number of visual units lies beween one and 45. In practice we see this optimum is between the number of phonemes and eight (which is the size of one of the smaller viseme sets). 

 For future work we are interested to extend these methods to work across speakers with a view to identify combinations of phonemes which can improve more than an single speaker.

  \eightpt
  \bibliographystyle{IEEEtran}

  \bibliography{mybib}

\end{document}